\documentclass{article}
\usepackage{spconf,amsmath,graphicx}
\usepackage{subfig}

\usepackage{multirow}
\usepackage{marvosym}

\renewcommand{\paragraph}[1]{\noindent\textbf{\textit{#1}}\;}

\title{SMCL: SALIENCY MASKED CONTRASTIVE LEARNING FOR\\LONG-TAILED VISUAL RECOGNITION}
\name{Sanglee Park$^1$ \quad Seung-won Hwang$^2$ \quad Jungmin So$^{1\dagger}$\thanks{$^{\dagger}$corresponding author (jso1@sogang.ac.kr). This work was supported in part by the NRF of Korea under grant no. 2022R1H1A2007390.}}
\address{$^1$Sogang University, Seoul, Republic of Korea\\
        $^2$Seoul National University, Seoul, Republic of Korea}
\begin{document}
%
\maketitle
\begin{abstract}
Real-world data often follow a long-tailed distribution with a high imbalance in the number of samples between classes. The problem with training from imbalanced data is that some background features, common to all classes, can be unobserved in classes with scarce samples. As a result, this background correlates to biased predictions into ``major" classes. In this paper, we propose saliency masked contrastive learning, a new method that uses saliency masking and contrastive learning to mitigate the problem and improve the generalizability of a model. Our key idea is to mask the important part of an image using saliency detection and use contrastive learning to move the masked image towards minor classes in the feature space, so that background features present in the masked image are no longer correlated with the original class. Experiment results show that our method achieves state-of-the-art level performance on benchmark long-tailed datasets.

\end{abstract}
\section{Introduction}
\label{sec:intro}

\begin{figure}[t]  
\centering
	\subfloat[major (bird)] {\label{fig:gradcam_major}\includegraphics[width=0.178\textwidth]{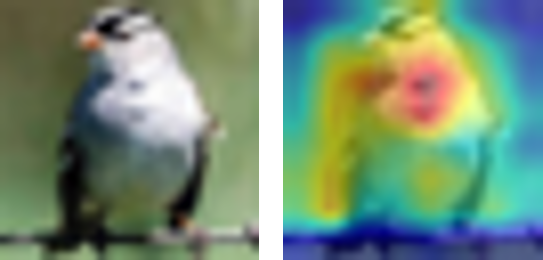}} \quad\,
	\subfloat[minor (horse)] {\label{fig:gradcam_horse}\includegraphics[width=0.27\textwidth]{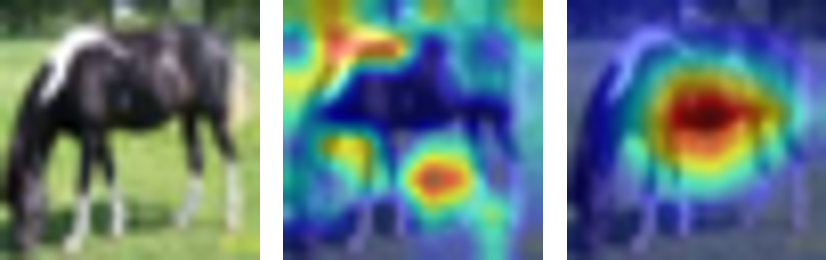}}
	\\
	\subfloat[saliency masked contrastive learning.] {\label{fig:concept}\includegraphics[width=0.42\textwidth]{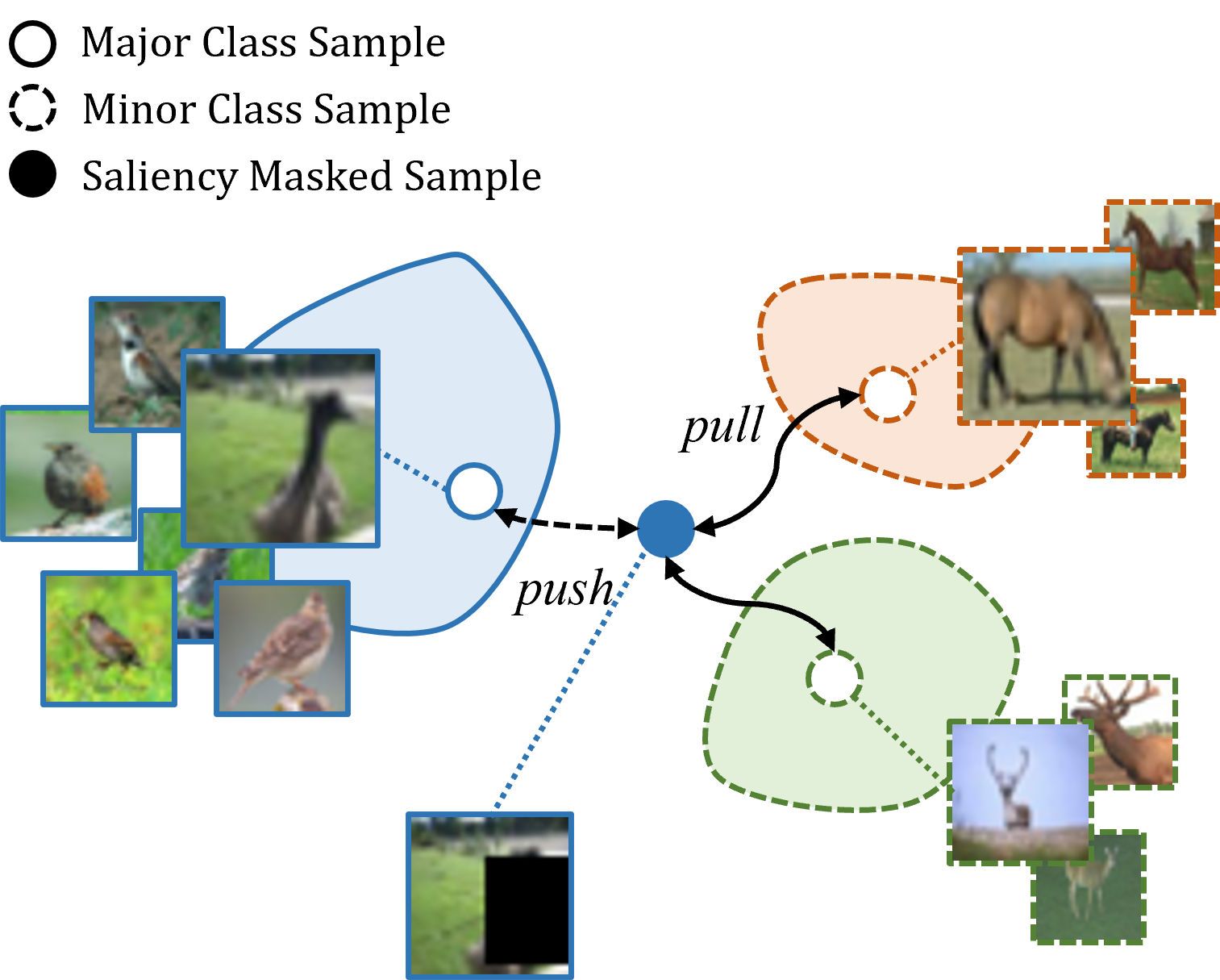}}
  \caption{(a): The image and its CAM of a major class sample (bird). (b): The image and its CAMs of a minor class sample (horse), misclassified as ``bird'' by the classifier. The middle image is the CAM for the predicted label (bird), and the right image is the CAM for the true label (horse). (c): Illustration of saliency masked contrastive learning.}
  \label{fig:idea}
\end{figure}

Deep neural networks have shown remarkable performance across many computer vision tasks such as image recognition and object detection. It is well known that outstanding performance results from training with a large amount of labeled data. However, many real-world datasets follow a long-tailed distribution, where the number of samples in each class differs greatly. In these datasets, the ``major'' classes have a large number of samples while the ``minor'' classes have only a few samples in each class. When a model is trained with a long-tailed dataset, the imbalance between classes hinders the model from correctly learning discriminative features.

We focus on the problem where a background feature that should be common to all classes is mistakenly correlated with a particular class. While this problem also occurs in balanced datasets, its effect is amplified in long-tailed datasets due to the lack of diversity in minor classes. For example, background features such as seas and trees should not affect class prediction, but may not be observed in minor classes, but only in some major classes. This biased observation mistakenly biases some background to a major class prediction. 

Examples in Fig.\ref{fig:gradcam_major} and \ref{fig:gradcam_horse} show the ``biased background'' problem of a model trained from imbalanced data. The image of a horse (a minor class) in Fig.\ref{fig:gradcam_horse} is misclassified as a bird (a major class) by the model. Fig.\ref{fig:gradcam_horse} shows two class activation maps (CAMs) obtained from the model. The left CAM is associated with the predicted label ``bird'', in which the grass area is highlighted. On the other hand, the right CAM is associated with the true label ``horse'', where the body of the animal is given focus. In other words, the model thinks the body looks like a horse, but the grass part is a clue that this image is a bird. Since the background feature (grass) is strongly attached to the bird class, the model outputs ``bird'' as the final decision. The bias is also observed in the CAM image of a bird in Fig.\ref{fig:gradcam_major}, in which the green area around the bird is highlighted, although the area is not related to the bird itself.

Our goal is to move the representation of background features presented in major classes towards minor classes in feature space so that the features are shared with minor classes and no longer biased towards major classes. To this end, we propose a new method called \textit{saliency masked contrastive learning} (SMCL). As illustrated in Fig.\ref{fig:concept}, the idea is to mask the salient part of an image so that only the background part remains, and use contrastive learning to pull the masked sample towards minor classes in feature space. While the source image is selected from the original long-tailed distribution, the target class where the masked image moves to is selected from a distribution that gives higher weights to minor classes. 

Previous methods exist where data augmentation is used to transfer the rich context features from major classes to minor classes \cite{park2021cmo}. Our key difference is that we use contrastive learning with masked data in order to move background features towards minor classes in feature space, compared to previous practices where augmented data are assigned labels and trained with cross-entropy-only loss. As discussed in experiments and ablation studies, applying contrastive learning is more effective than training them with cross-entropy-only loss. Experiment results show that the proposed method achieves comparable performance with the current state-of-the-art on the benchmark long-tailed datasets.



\section{Related Work}
\label{sec:relatedwork}

\paragraph{Long-tailed Recognition.}
Long-tailed recognition is an important task since many practical datasets are imbalanced. Initial approaches for long-tailed learning include re-sampling \cite{park2021cmo} where more samples are selected from minor classes to achieve a balance between classes, and re-weighting \cite{cui2019class} where the loss function is adjusted to give more weight to minor class losses. Many variations of re-sampling and re-weighting techniques have been proposed \cite{cao2019learning,Kang2020Decoupling,ren2020balanced,zhong2021improving}, but they do not address the problem of biased background features. Mixup methods \cite{Yun_2019_ICCV,uddin2021saliencymix,park2021cmo,zhang2018mixup,chou2020remix} can help improve long-tailed learning by mixing two images of different classes and assigning mixed labels. However, they do not apply contrastive learning with mixed samples in order to obtain a better representation.

\paragraph{Supervised Contrastive Learning.}
Supervised contrastive learning (SCL) \cite{chen2020simple, khosla2020supervised} pulls together samples belonging to the same class and pushes apart samples from different classes in the feature space. Recently, SCL is applied to long-tailed learning to maintain uniform distribution in the feature space and improve class boundaries \cite{wang2021contrastive, li2022targeted}. For example, TSC \cite{li2022targeted} generates uniformly distributed targets on a hypersphere and uses contrastive learning to make the features of different classes converge to the targets. Different from previous schemes, our proposed method uses supervised contrastive learning so that the masked data containing background features are pulled towards minor class samples in the feature space.


\section{Proposed Method}
\label{sec:method}

\begin{figure}[tb]

\centering
\centerline{\includegraphics[width=0.37\textwidth]{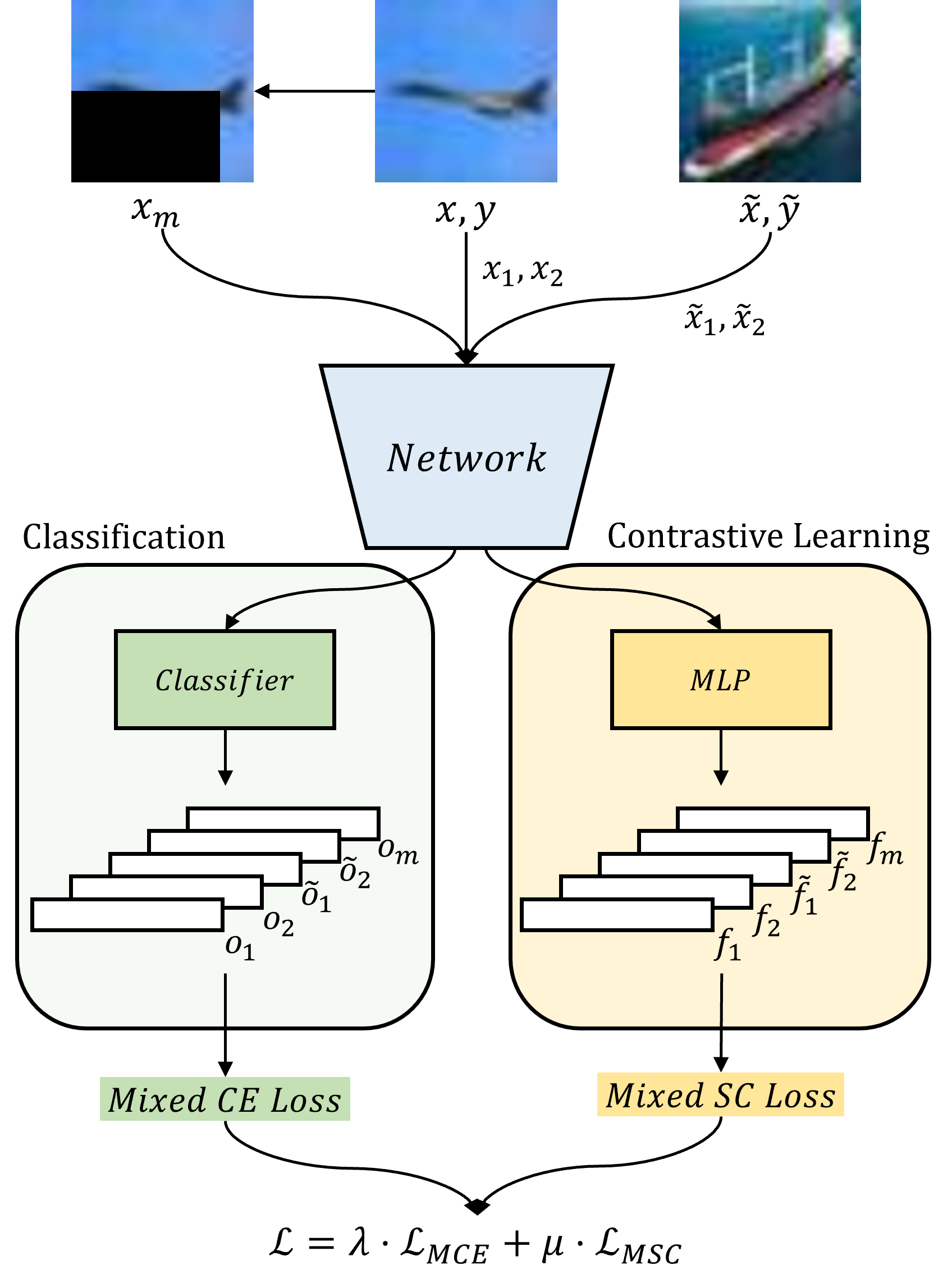}}
%
\caption{Overview of the proposed framework.}
\label{fig:framework}
\end{figure}

Overview of the proposed method SMCL is shown in Fig.\ref{fig:framework}. SMCL involves saliency masking to produce background images, weighted sampling to select target labels giving priority to minor classes, and saliency masked contrastive learning to pull the background image towards minor classes in the feature space.

\subsection{Saliency Masking}
\label{ssec:saliency-masking}

In order to mask discriminative features that belong to a certain class, we use saliency detection methods. Similarly \cite{uddin2021saliencymix}, we use the saliency function $S(\cdot)$ and find the pixel $i$, $j$ that has the maximum value:
\begin{equation}\label{eq:1}
    \begin{array}{cc}
        i, j = \mathrm{argmax}(S(x)),
    \end{array}
\end{equation}
where $x$ is a sample in the training set $\mathcal{D}$. Then, following the convention \cite{Yun_2019_ICCV}, we mask a region of the size generated following a beta distribution $\mathrm{Beta}(\alpha, \alpha)$ centered at ($i$, $j$).




\subsection{Weighted Sampling}
\label{ssec:weighted-sampling}

For each data in the training set, we sample target data to be used in masked contrastive learning. In order to increase the chance of selecting a minor class label, we follow the sampling strategy \cite{park2021cmo} based on the effective number calculated from class size \cite{cui2019class}:
\begin{equation}\label{eq:2}
    E_{k} = \frac{(1-\beta^{n_k})}{(1-\beta)},
\end{equation}
where $n_k$ is the number of samples in $k$th class, $N=\sum_{k}{n_k}$, and $\beta = (N-1)/N$. Based on $E_k$, the sampling probability of class $k$ is:
\begin{equation}\label{eq:3}
    p_k = \frac{1/E_k}{\sum_{k}{1/E_k}},
\end{equation}
which leads to a higher sampling probability for the minor classes. The ``minor weighted'' distribution $\tilde{\mathcal{D}}$ is a distribution resulting from sampling based on $p_k$.

\subsection{Saliency Masked Contrastive Learning}
\label{ssec:framework}

The key idea of SMCL is to transfer a background feature from a major class to a minor class. For an image $x$ in a batch, we sample a target image $\tilde{x}$ from the minor weighted distribution $\tilde{\mathcal{D}}$. Then we use basic augmentation schemes such as random crop and random flip to create two versions of each sample, $x_1$, $x_2$, $\tilde{x_1}$, $\tilde{x_2}$. Finally, we use saliency masking to generate the masked image $x_m$. Inserting the five samples into the model, we obtain their logits $O$ and feature vectors $F$
\begin{equation}\label{eq:4}
    \begin{array}{cc}
    O, F = J([x,\tilde{x},x_m]),
    \end{array}
\end{equation}
where $O=[o_1, o_2, \tilde{o}_1, \tilde{o}_2, o_m]$ and $F=[f_1, f_2, \tilde{f}_1, \tilde{f}_2, f_m]$. For the loss function, we use a combination of the cross-entropy loss and the contrastive loss. The cross-entropy loss $\mathcal{L}_{MCE}$ is calculated as:
\begin{equation}\label{eq:5}
    \mathcal{L}_{MCE} = (1-A)\cdot\mathcal{L}_{CE}(o, y)+A\cdot\mathcal{L}_{CE}(\tilde{o}, \tilde{y}),
\end{equation}
where $o=[o_1, o_2, o_m]$, $\tilde{o}=[\tilde{o}_1, \tilde{o}_2, o_m]$, $\mathcal{L}_{CE}(o, y)$ is the cross-entropy between the logits $o$ and their true labels $y$, and $A$ indicates the proportion of the masked region. Next, we define the supervised contrastive loss $\mathcal{L}_{SC}$ as:
\begin{equation}\label{eq:6}
    \mathcal{L}_{SC} = -\frac{1}{|B_y|-1} \sum_{p\in B_y \setminus \{z\} }\log{\frac{\exp(f_z\cdot f_p/\tau)}{\sum_{k\in B\setminus \{z\}} \exp(f_z\cdot f_k/\tau)}},
\end{equation}
where $B$ is the batch, $B_y$ is the set of samples in the batch that has the label $y$, $z$ is the source image, $f$ is the feature vector, and $\tau$ is a temperature hyperparameter. We use a mixed contrastive loss $\mathcal{L}_{MSC}$:
\begin{equation}\label{eq:7}
    \mathcal{L}_{MSC} = (1-A)\cdot\mathcal{L}_{SC}(f, y)+A\cdot\mathcal{L}_{SC}(\tilde{f}, \tilde{y}),
\end{equation}
where $f=[f_1, f_2, f^m]$, $\tilde{f}=[\tilde{f}_1, \tilde{f}_2, f^m]$, and $\mathcal{L}_{SC}(f, y)$ is the supervised contrastive loss when $f$ is the set of feature vectors and $y$ is the set of true labels associated with the feature vectors. Finally, our loss function combines the cross-entropy loss and the contrastive loss, where $\lambda$ and $\mu$ are hyperparameters.
\begin{equation}\label{eq:8}
    \mathcal{L} = \lambda \cdot \mathcal{L}_{MCE} + \mu \cdot \mathcal{L}_{MSC}.
\end{equation}


\section{Experiments}
\label{sec:experiments}

\subsection{Datasets}
\label{ssec:datasets}

\paragraph{CIFAR-10-LT and CIFAR-100-LT.}
CIFAR-LT datasets are constructed from CIFAR \cite{krizhevsky2009learning} by adjusting the dataset into a long-tailed shape. In our evaluations, we consider three different imbalance ratios $\rho \in \{10, 50, 100\}$ for CIFAR-10 and CIFAR-100 respectively.

\paragraph{ImageNet-LT.} ImageNet-LT is generated by selecting the long-tailed subset from ImageNet-2012 \cite{ILSVRC15}. The imbalance ratio $\rho$ is 256 and has 115.8K images from 1,000 categories.

\subsection{Implementation Settings}
\label{ssec:implement}

\paragraph{CIFAR-10-LT and CIFAR-100-LT.} We train ResNet-32 \cite{he2016deep} for 200 epochs using stochastic gradient descent (SGD) with a batch size of 256, a momentum of 0.9, and a weight decay of 2e-4. The initial learning rate is 0.1 and we decay the learning rate at epochs 160 and 180 by 0.1 \cite{cao2019learning}. We use CutOut \cite{cutout} and SimAugment \cite{chen2020simple} as augmentation strategies. saliency masking is applied with a probability of 0.2 starting from epoch 160. The hyperparameters $\lambda$ and $\mu$ are empirically tuned to $\lambda=1.0$ and $\mu=0.3$.

\paragraph{ImageNet-LT.} We train ResNet-50 for 100 epochs using SGD with a batch size of 256, a momentum of 0.9, and a weight decay of 5e-4. The initial learning rate is set to 0.1 and follows cosine scheduling. We used RandAug and SimAugment as the augmentation strategies. Saliency masking is applied with a probability of 0.4. Both augmentation and re-weighting are applied from epoch 80 \cite{cao2019learning}. The hyperparameters $\lambda$ and $\mu$ are empirically tuned to $\lambda=1.0$ and $\mu=0.35$.

The results from Mixup and CutMix are obtained following the same setup from CMO \cite{park2021cmo}. 

\subsection{Experimental Results}
\label{ssec:results}

We compare the performance of SMCL with previous methods \cite{cao2019learning, cui2019class, zhou2020bbn, zhong2021improving, cui2022reslt}, including mixup augmentation based methods \cite{zhang2018mixup, Yun_2019_ICCV, chou2020remix, park2021cmo}, and supervised contrastive learning based methods \cite{khosla2020supervised, li2022targeted, wang2021contrastive}.

Results on CIFAR-10/100-LT using different imbalance ratios are shown in Table \ref{tab:cifar-acc}. SMCL outperforms previous methods on CIFAR-10/100-LT when applied with deferred re-weighting (DRW) \cite{cao2019learning}. Compared with baseline ERM and Supervised Contrastive Learning (SCL), SMCL achieves 6.2\% and 5.1\% improvements in CIFAR-100-LT (100), respectively. Furthermore, SMCL significantly outperforms similar augmentation methods, such as Remix \cite{chou2020remix} and CMO \cite{park2021cmo}. SMCL also shows consistent improvements even when compared with SCL based methods.

\begin{table}[tb]
\centering
\resizebox{\columnwidth}{!}{%
\begin{tabular}{l|ccc|ccc}
\hline
\multicolumn{1}{c|}{Dataset} & \multicolumn{3}{c|}{CIFAR-100-LT}                & \multicolumn{3}{c}{CIFAR-10-LT}                  \\ \hline
\multicolumn{1}{c|}{Imbalance Ratio}                        & 100          & 50            & 10            & 100           & 50            & 10            \\ \hline
ERM\textsuperscript{\Cross}                                 & 38.3         & 43.9          & 55.7          & 70.4          & 74.8          & 86.4          \\
DRW\textsuperscript{\Cross}                                 & 41.5         & 45.3          & 58.1          & 76.3          & 80.0          & 87.6          \\
LDAM-DRW\textsuperscript{\Cross} \cite{cao2019learning}     & 42.0         & 46.6          & 58.7          & 77.0          & 81.0          & 88.2          \\
CB Focal \cite{cui2019class}                                & 39.6         & 45.3          & 58.0          & 74.6          & 79.3          & 87.5          \\
BBN\textsuperscript{\Cross} \cite{zhou2020bbn}              & 42.6         & 47.0          & 59.1          & 79.8          & 82.2          & 88.3          \\
MiSLAS\textsuperscript{$\ast$} \cite{zhong2021improving}    & 47.0         & 52.3          & 63.2           & 82.1          & 85.7         & 90.0          \\
ResLT\textsuperscript{$\ast$} \cite{cui2022reslt}           & 48.2         & 52.7          & 62.0          & 82.4           & 85.2         & 89.7          \\ \hline
DRW+Mixup \cite{zhang2018mixup}                             & 45.3         & 49.7          & 60.3          & 81.1          & 83.8          & 89.1          \\
DRW+CutMix \cite{Yun_2019_ICCV}                             & 46.4         & 51.2          & 62.1          & 81.8          & 85.1          & 89.7          \\
ReMix\textsuperscript{$\ast$} \cite{chou2020remix}          & 46.8         & -             & 61.2          & 79.8          & -             & 89.0          \\
DRW+CMO \cite{park2021cmo}                                  & 46.4         & 51.5          & 61.7          & 81.7          & 84.9          & 89.6          \\ \hline
SCL                                                         & 39.4         & 44.4          & 56.7          & 72.2          & 76.8          & 87.1          \\
SCL-DRW                                                     & 41.7         & 46.8          & 57.9          & 75.4          & 79.3          & 87.9          \\
TSC\textsuperscript{$\ast$} \cite{li2022targeted}           & 43.8         & 47.4          & 59.0          & 79.7          & 82.9          & 88.7          \\
Hybrid-SC\textsuperscript{$\ast$} \cite{wang2021contrastive}& 46.7         & 51.9          & 63.1          & 81.4          & 85.4          & \textbf{91.1} \\ \hline\hline
SMCL (Ours)                                                 & 44.5         & 49.8          & 61.8          & 80.2          & 84.3          & 90.3          \\
DRW+SMCL (Ours)                                             & \textbf{50.1}& \textbf{54.8} & \textbf{64.1} & \textbf{84.2} & \textbf{86.8} & 91.0 \\ \hline
\end{tabular}%
}
\caption{Comparison of classification accuracy(\%) on the CIFAR-10/100-LT datasets using ResNet-32. $\ast$ and \Cross\, are results from the original papers and \cite{zhou2020bbn}, respectively.}
\label{tab:cifar-acc}
\end{table}

\begin{table}[tb]
\centering
\resizebox{0.42\textwidth}{!}{%
\begin{tabular}{l|rrrr}
\hline
Methods                                                         & All  & Many & Med  & Few   \\ \hline
ERM\textsuperscript{\Cross}                                     & 41.6 & 64.0 & 33.8 &  5.8 \\
Decouple-cRT\textsuperscript{\Cross} \cite{Kang2020Decoupling}  & 47.3 & 58.8 & 44.0 & 26.1 \\
Decouple-LWS\textsuperscript{\Cross} \cite{Kang2020Decoupling}  & 47.7 & 57.1 & 45.2 & 29.3 \\
DRW                                                             & 49.9 & 61.2 & 47.1 & 29.2 \\
LDAM-DRW                                                        & 50.3 & 61.7 & 47.0 & 30.6 \\
Balanced Softmax \cite{ren2020balanced}                         & 50.6 & 60.1 & 48.4 & 32.6 \\ \hline
DRW+Mixup                                                       & 49.3 & 60.7 & 46.4 & 28.7 \\
DRW+CutMix                                                      & 50.3 & 60.7 & 47.6 & 31.7 \\
DRW+CMO                                                         & 50.8 & 61.2 & 48.7 & 30.1 \\ \hline
SCL                                                             & 45.9 & 69.0 & 38.7 &  8.2 \\
SCL-DRW                                                         & 49.4 & 61.6 & 47.2 & 24.1 \\
TSC\textsuperscript{$\ast$}                                     & \textbf{52.4} & 63.5 & \textbf{49.7} & 30.4 \\ \hline\hline
SMCL (Ours)                                                     & 46.0 & \textbf{67.9} & 39.1 & 10.3 \\
DRW+SMCL (Ours)                                                 & 51.4 & 60.4 & \textbf{49.7} & \textbf{32.8} \\ \hline
\end{tabular}
}
\caption{Comparison of classification accuracy(\%) on the ImageNet-LT dataset using ResNet-50. $\ast$ and \Cross\, are results from the original papers and \cite{Kang2020Decoupling}, respectively.}
\label{tab:imagenet}
\end{table}

Results for ImageNet-LT are reported in Table \ref{tab:imagenet}. The table shows the accuracy of three subsets: Many shots ($>$100 samples), medium shots (20-100 samples), and few shots ($<$20 samples). SMCL shows comparable performance with the state-of-the-art \cite{li2022targeted}, showing 2.4\% better performance on a few shots. Since ImageNet-LT is composed of 38.5\% of many, 46.9\% of medium, and 14.6\% of few classes, improved performance in few has relatively little impact on overall performance. However, it is noteworthy that applying SMCL with DRW outperforms most of the existing methods, especially achieving considerable gains in the accuracy of a few shots.


\subsection{Ablation Study}
\label{ssec:ablation}

\paragraph{Impact of Saliency Masking.}
To show the impact of saliency masking, we tried masking images with two other methods: masking random area and masking the center. As shown in Table \ref{tab:saliency}, covering the center of an image achieved 0.3\% higher than the random. Since many images have their core feature at the center, masking the center area is more useful than masking a random area. Saliency masking showed the highest result, approximately 0.3\% higher than center masking. This result shows that saliency masking is indeed effective in selecting background features from an image.
\\

\begin{table}[htb]
\centering
\resizebox{0.33\textwidth}{!}{%
\begin{tabular}{l|ccc}
\hline
          & Random & Center & Saliency       \\ \hline
Acc. (\%) & 49.56  & 49.82  & \textbf{50.14} \\ \hline
\end{tabular}
}
\caption{Performance of saliency masking compared with other masking methods on CIFAR-100-LT ($\rho=100$).}
\label{tab:saliency}
\end{table}

\paragraph{Effect of Contrastive Learning.}
To evaluate the effect of saliency masked contrastive learning, we compared the performance of models with and without contrastive learning. In Table \ref{tab:scl-loss}, ``Cross Entropy'' indicates the accuracy when the model is trained with only the cross-entropy loss. The results show that applying contrastive learning is effective in improving model performance.


\begin{table}[htb]
\centering
\resizebox{0.35\textwidth}{!}{%
\begin{tabular}{c|cc}
\hline
\multicolumn{1}{l|}{}                                                          & SMCL           & DRW+SMCL       \\ \hline
Cross Entropy                                                                  & 43.57          & 49.12          \\ \hline
\begin{tabular}[c]{@{}c@{}}Supervised\\Contrastive Learning\end{tabular}       & \textbf{44.49} & \textbf{50.14} \\ \hline
\end{tabular}
}
\caption{Performance of applying saliency masked contrastive learning compared with cross-entropy-only learning on CIFAR-100-LT ($\rho=100$).}
\label{tab:scl-loss}
\end{table}

\section{Conclusion}
\label{sec:conclusion}
We proposed SMCL, a saliency masked contrastive learning method that achieves high accuracy on long-tailed datasets. SMCL uses saliency masking to obtain background features from images and applies contrastive learning to move their embeddings towards minor classes so that they are detached from major classes. The proposed method is simple to implement, yet achieves state-of-the-art level performance on benchmark long-tailed datasets such as CIFAR-10/100-LT and ImageNet-LT.


\vfill\pagebreak

\bibliographystyle{IEEEbib}
\bibliography{strings,refs}

\end{document}